\documentclass{article}

\usepackage[english]{babel}

\usepackage[letterpaper,top=2cm,bottom=2cm,left=3cm,right=3cm,marginparwidth=1.75cm]{geometry}

\usepackage{booktabs}

\usepackage[accsupp]{axessibility}  
\usepackage[ruled,vlined]{algorithm2e} 
\usepackage[normalem]{ulem}
\useunder{\uline}{\ul}{}

\usepackage{booktabs} 
\usepackage{multirow} 
\usepackage{caption}  
\usepackage{array}    
\usepackage{tabularx}

\usepackage{siunitx}     
\usepackage[table]{xcolor} 
 
\newcolumntype{C}{>{\centering\arraybackslash}X}

\usepackage{amsmath}
\usepackage{graphicx}
\usepackage[colorlinks=true, allcolors=blue]{hyperref}

\title{PromptForge-350k: A Large-Scale Dataset and Contrastive Framework for Prompt-Based AI Image Forgery Localization}
\author{Jianpeng Wang, Haoyu Wang, Baoying Chen, \\Jishen Zeng, Yiming Qin, Yiqi Yang, Zhongjie Ba}

\begin{document}
\date{}
\maketitle

\begin{abstract}
The rapid democratization of prompt-based AI image editing has recently exacerbated the risks associated with malicious content fabrication and misinformation. However, forgery localization methods targeting these emerging editing techniques remain significantly under-explored. 
To bridge this gap, we first introduce a fully automated mask annotating framework that leverages keypoint alignment and semantic space similarity to generate precise ground-truth masks for edited regions. Based on this framework, we construct PromptForge-350k, a large-scale forgery localization dataset covering four state-of-the-art prompt-based AI image editing models, thereby mitigating the data scarcity in this domain.
Furthermore, we propose ICL-Net, an effective forgery localization network featuring a triple-stream backbone and intra-image contrastive learning. This design enables the model to capture highly robust and generalizable forensic features.
Extensive experiments demonstrate that our method achieves an IoU of 62.5\% on PromptForge-350k, outperforming SOTA methods by 5.1\%. Additionally, it exhibits strong robustness against common degradations with an IoU drop of less than 1\%, and shows promising generalization capabilities on unseen editing models, achieving an average IoU of 41.5\%.

\end{abstract}

\section{Introduction}
\label{sec:intro}

\begin{figure}[t]
    \centering
    \includegraphics[width=0.8\linewidth]{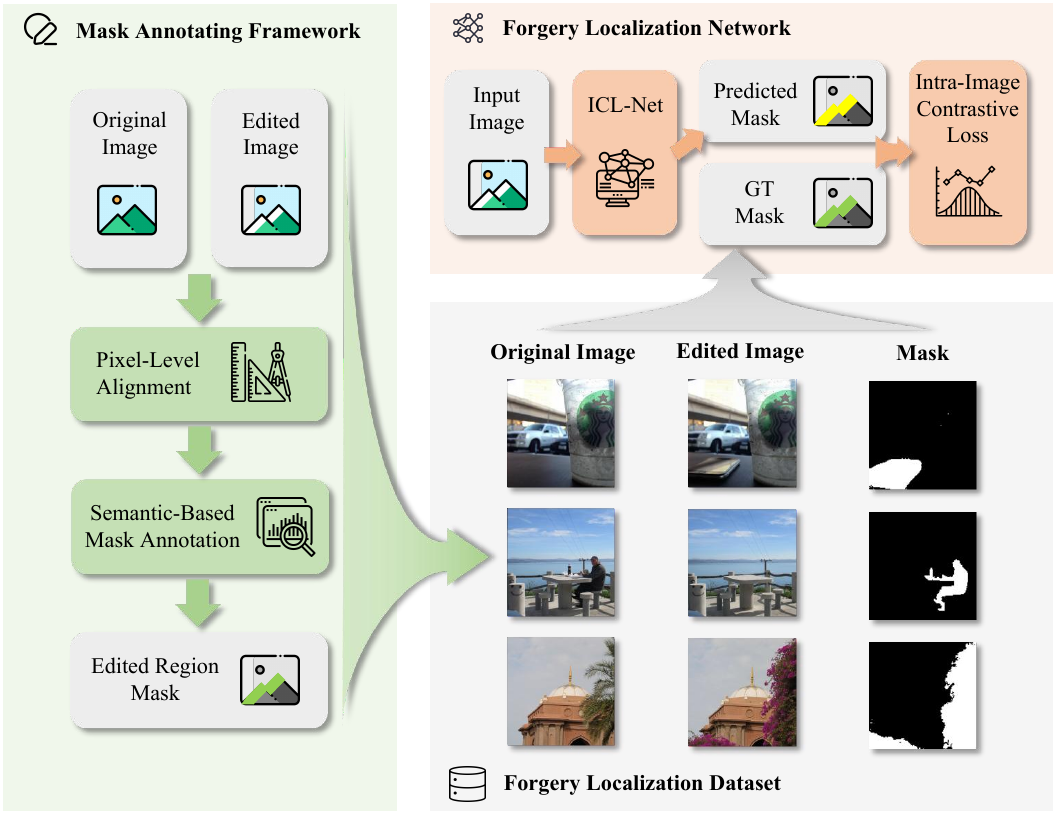}
    \caption{Overview of our work: (1) A fully automated mask annotating framework targeting prompt-based AI image editing. (2) PromptForge-350k, a comprehensive forgery localization dataset. (3) ICL-Net, an effective forgery localization network.}
    \label{fig:introduction}
\end{figure}



The rapid advancement of Generative AI technologies has made AI-powered image editing tools increasingly accessible and ubiquitous. By leveraging prompt-based image editing models, such as Flux.Kontext \cite{labsFLUX1KontextFlow2025a} and Nano-Banana \cite{comaniciGemini25Pushing2025}, users can generate coherent manipulated images by merely providing an input image and an editing instruction. Nevertheless, this lowered barrier to image manipulation has facilitated the widespread dissemination of forged content online, exacerbating risks associated with misinformation and fraud. Consequently, developing forgery localization methods tailored to these emerging image editing techniques has become critical.

Despite this urgency, current research on image forgery localization lags significantly behind the rapid evolution of AI-powered image editing technologies.

In terms of \textbf{forgery localization datasets}, existing datasets are primarily limited to traditional Photoshop \cite{guan2019mfc,mahfoudi2019defacto,novozamskyIMD2020LargeScaleAnnotated2020} or mask-based AI editing \cite{caiZoomingFakesNovel2025a,chenGIMMillionscaleBenchmark,mareenTGIFTextGuidedInpainting2024}, failing to cover emerging prompt-based techniques. Furthermore, since prompt-based models regenerate the entire image without explicit edited region masks, the absence of automated annotation methodologies poses a significant hurdle.

Regarding \textbf{forgery localization methods}, existing approaches rely heavily on low-level artifacts such as splicing traces\cite{dongMVSSNetMultiViewMultiScale2023}, sensor noise\cite{baiPIMNetProgressiveInconsistency2025a}, and JPEG compression inconsistencies\cite{zhuMesoscopicInsightsOrchestrating}. However, these subtle forensic traces are often disrupted or erased during the generative reconstruction process inherent to prompt-based editing. This necessitates the exploration of localization strategies that look beyond low-level signal patterns.


In this work, as illustrated in Fig \ref{fig:introduction},  we first introduce an automated \textbf{mask annotation framework} based on pixel-level alignment and semantic space feature similarity. This approach guarantees the accuracy of edited region mask annotations while eliminating the need for labor-intensive manual labeling. Building upon this framework, we construct PromptForge-350k, a comprehensive \textbf{forgery localization dataset} comprising 354,258 edited image pairs annotated with precise pixel-level edited region masks. This dataset encompasses diverse samples generated by four state-of-the-art open-source or closed-source image editing models (Nano-banana\cite{comaniciGemini25Pushing2025}, BAGEL\cite{dengEmergingPropertiesUnified2025b}, Flux.Kontext\cite{labsFLUX1KontextFlow2025a}, Step1x-edit\cite{liuStep1XEditPracticalFramework2025a}) across eight distinct editing tasks. Furthermore, we propose ICL-Net, a specialized \textbf{forgery localization network} that exploits \textbf{I}ntra-image \textbf{C}ontrastive \textbf{L}earning to effectively capture forensic traces in prompt-based AI edited images. Our contributions are summarized as follows:

\begin{itemize}
    \item  We introduce a fully automated framework capable of generating precise edited-region masks for prompt-based AI forged images, circumventing the need for manual annotation.
    
    \item  We construct a comprehensive forgery localization dataset, PromptForge-350k, containing automatically annotated ground-truth masks, mitigating the data scarcity in this domain.
    
    \item We propose ICL-Net, an effective forgery localization network driven by intra-image contrastive learning to capture robust and generalizable forensic features.
\end{itemize}

Quantitative results demonstrate the effectiveness of our approach, which achieves an IoU of 62.5\% on the PromptForge-350k dataset, surpassing existing SOTA methods by 5.1\%. Notably, ICL-Net demonstrates high robustness against common image degradations, exhibiting a negligible IoU drop of less than 1\%. It also generalizes well to unseen editing models, achieving an average IoU of 41.5\%. Both the dataset and source code will be made publicly available to facilitate reproducibility and future research.

\section{Related Works}
\label{sec:related_works}

\subsection{AI Image Forgery Datasets}


Existing AI image forgery datasets can be categorized into two types based on the paradigms of the image editing methods used for their construction. Early \textbf{mask-based editing} methods require users to manually draw a mask over the editing region and provide instructions; the model then regenerates the content within the masked area while preserving the remaining regions. Conversely, the emerging \textbf{prompt-based editing} paradigm eliminates the dependency on explicit edited region masks, allowing users to merely provide editing instructions via natural language. While mature dataset construction methodologies and large-scale datasets exist for mask-based approaches, forgery localization datasets tailored to prompt-based AI editing remain largely unexplored.

 \textbf{Mask-based Editing Dataset.} 
MagicBrush \cite{zhangMAGICBRUSHManuallyAnnotated} utilizes manually annotated masks to perform image editing via DALL-E 2. TGIF \cite{mareenTGIFTextGuidedInpainting2024} employes images and corresponding captions from MSCOCO as inputs for mask-based AI editing models. GRE \cite{sunRethinkingImageEditing2024a}, FakeShield \cite{xuFakeShieldExplainableImage2025}, and GIM \cite{chenGIMMillionscaleBenchmark} adopt a similar approach: employing the SAM model to generate object masks and LLMs to synthesize editing instructions. Additionally, BRGen \cite{caiZoomingFakesNovel2025a} addresses background editing scenarios often overlooked by existing datasets, thereby enhancing dataset diversity.

 \textbf{Prompt-based Editing Dataset.}
Several recent works have attempted to construct datasets tailored to prompt-based image editing. UltraEdit \cite{zhaoUltraEditInstructionbasedFineGrained} employs a prompt-to-prompt strategy to control SDXL for generating edited images. X-Edit \cite{bazylevaXEditDetectingLocalizing2025} utilizes InstructPix2Pix\cite{brooksInstructPix2PixLearningFollow2023} to generate edited images, obtaining approximate edited region through pixel differencing. Both PicoBanana \cite{qianPicoBanana400KLargeScaleDataset2025} and X2Edit \cite{maX2EditRevisitingArbitraryInstruction2025} constructed datasets using state-of-the-art prompt-based editing methods, but lack edited region masks.

\subsection{Image Forgery Localization Methods}
Image editing inevitably introduces pixel-level discrepancies between edited and non-edited regions, which serve as critical forensic cues. To exploit these discrepancies, the community has developed diverse forgery localization methodologies, which can be broadly categorized into three streams:

 \textbf{Feature-Centric Approaches.} Many works focus on high-frequency or statistical anomalies. MVSS-Net~\cite{dongMVSSNetMultiViewMultiScale2023}, PSCC-Net~\cite{liuPSCCNetProgressiveSpatioChannel2022}, and HiFiNet~\cite{guoHierarchicalFineGrainedImage2023} incorporate noise-sensitive branches, dual-path structures, and LoG filtering respectively to extract pixel-level artifacts. Similarly, PIM-Net~\cite{baiPIMNetProgressiveInconsistency2025a} leverage CMOS noise patterns in non-edited regions to identify forged areas.

 \textbf{Model Architecture Designs.} Recent studies have introduced specialized architectures to extract forgery features. APSC-Net~\cite{quModernImageManipulation2024a} employs multi-scale extractors for holistic analysis. SparseViT~\cite{suCanWeGet2025} performs attention within grouped feature maps, a concept further extended by NFA-ViT~\cite{caiZoomingFakesNovel2025a} via noise-guided amplification, to mitigate semantic interference. Additionally, Mesoscopic Insights~\cite{zhuMesoscopicInsightsOrchestrating} combines parallel CNN-Transformer encoders with DCT enhancement to capture mesoscopic features.

 \textbf{Advanced Learning Paradigms.} Beyond architectural innovations, advanced learning strategies have been adopted to enhance forgery localization efficacy. TruFor~\cite{guillaroTruForLeveragingAllround2023}, NCL-IML~\cite{zhouPretrainingfreeImageManipulation2023}, and FOCAL~\cite{wuRethinkingImageForgery2025} harness contrastive learning, while CoDE~\cite{pengEmployingReinforcementLearning2024} and AdaIFL~\cite{liAdaIFLAdaptiveImage2025a} integrate reinforcement learning and dynamic routing. Furthermore, DiffForensics~\cite{yuDiffForensicsLeveragingDiffusion2024} and InPDiffusion~\cite{wangInpDiffusionImageInpainting2025} reformulate localization as a mask generation task using conditional diffusion models.


\section{Mask Annotating Framework}
\label{sec:framework}


In this section, we introduce the mask annotating framework employed for our PromptForge-350k dataset.
Building upon the original and edited images from the X2Edit\cite{maX2EditRevisitingArbitraryInstruction2025} and PicoBanana\cite{qianPicoBanana400KLargeScaleDataset2025} datasets, we select samples associated with local editing tasks to undergo the following annotation process: 

\textbf{Pixel-Level Alignment.}
We first perform pixel-level alignment between the original and edited images. This step is essential because certain image editing models (e.g., Nano-Banana and Kontext) internally apply pixel cropping to input images, which would cause spatial misalignment and compromise the subsequent mask annotation process.

\textbf{Semantic-Based Mask Annotation.}
Subsequently, we leverage DINO v3 \cite{voDINOv32025} to extract dense semantic features from the aligned images and compute the cosine similarity at corresponding spatial positions. This strategy effectively circumvents pixel-level variances introduced by prompt-based methods in non-edited regions, enabling robust mask annotation.

\subsection{Pixel-Level Alignment}


\begin{figure}[htbp]
    \centering
    \includegraphics[width=1.0\linewidth]{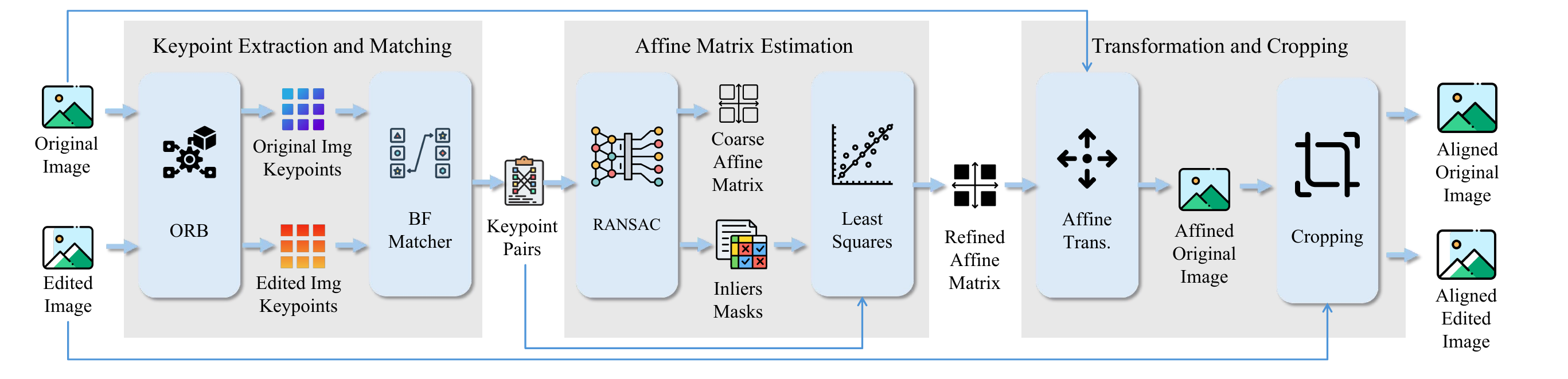}
    \caption{Pipeline of Pixel-Level Alignment: The process involves keypoint extraction and matching, followed by affine matrix estimation. Finally, we apply a coordinate transformation, and the black borders are cropped out.}
    \label{fig:alignment_pipeline}
\end{figure}

The alignment process comprises three stages: \textbf{keypoint extraction and matching, affine matrix estimation, transformation and cropping}, as illustrated in Fig \ref{fig:alignment_pipeline}. For implementation details, we provide the pseudo code in Algorithm 1 of the supplementary material.

 Formally, as defined in Equation~\ref{eq:affine_2d}, we employ a 6-degree-of-freedom 2D affine to model geometric distortions, including scaling, translation, and shearing. The parameters $a_1,\dots,a_6$ denote the transformation coefficients to be estimated.
 
\begin{equation}
\begin{bmatrix}
x' \\
y' \\
1
\end{bmatrix}
=
\underbrace{
\begin{bmatrix}
a_1 & a_2 & a_3 \\
a_4 & a_5 & a_6 \\
0 & 0 & 1
\end{bmatrix}
}_{\text{transformation matrix } \mathbf{A}}
\begin{bmatrix}
x \\
y \\
1
\end{bmatrix}
\label{eq:affine_2d}
\end{equation}

The transformation matrix $A$ defines the coordinate mapping between the original and edited images. Therefore, the primary objective of our pixel alignment process is to solve for this matrix $A$.

\subsubsection{Keypoint Extraction and Matching.} First, we employ the ORB algorithm~\cite{rubleeORBEfficientAlternative2011} to extract 1,000 keypoints from the original and edited images, respectively. Each keypoint is characterized by its spatial coordinate and a 32-byte binary descriptor encoding local features. Keypoint matching between the two groups of keypoints is subsequently performed using a Brute-Force Matcher (BFMatcher) based on the Hamming distance. To ensure match reliability, we apply Lowe’s ratio test, discarding matches where the distance ratio between the nearest and second-nearest matching options exceeds 0.75.

\subsubsection{Affine Matrix Estimation.} Next, to handle potential noise and outliers in the matched keypoint pairs, we utilize the RANSAC~\cite{fischlerRandomSampleConsensus} algorithm. This iterative algorithm robustly estimates the model parameters supported by the maximum number of inliers. Specifically, applying RANSAC to the keypoint pairs yields a coarse transformation matrix A and a set of inlier indicators, denoted as $idx_{inliers}$. This boolean list indicates whether a corresponding keypoint pair is consistent with the coarse matrix A (i.e., is an inlier). Leveraging $idx_{inliers}$, we select the inlier keypoint pairs and compute a refined affine transformation matrix, $A_{refined}$, by solving a least-squares problem on this inlier subset.

\subsubsection{Transformation and Cropping.} Once $A_{refined}$ is obtained, we warp the original image to align it with the coordinate system of the edited image. Since affine transformations may introduce black borders at the image boundaries, we apply an identical cropping operation to both the transformed original image and the edited image to eliminate such regions.

\subsection{Semantic-based Mask Annotation}

\begin{figure}[t]
    \centering
    \includegraphics[width=0.8\linewidth]{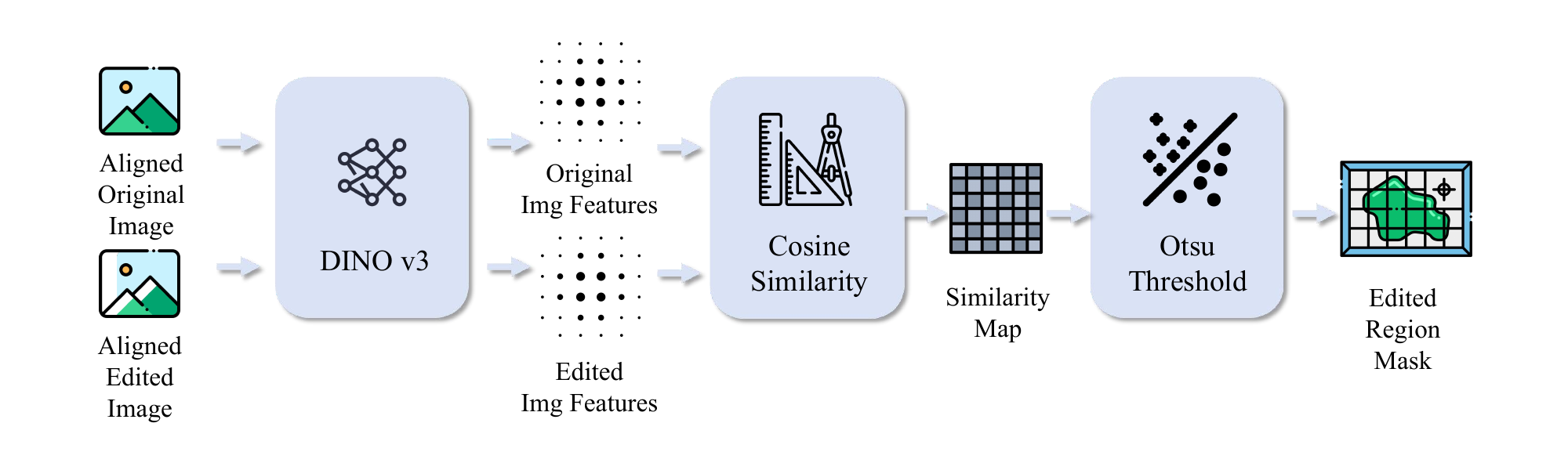}
    \caption{Pipeline of Semantic-Based Mask Annotation. We first utilize DINO v3 to extract semantic features, followed by calculating the pixel-wise feature similarity. Finally, the similarity map is binarized to output the edited region mask.}
    \label{fig:feature_extraction}
\end{figure}

Prompt-based editing models typically induce global pixel-level changes, rendering naive pixel-level differencing ineffective for accurate mask estimation. Consequently, we propose utilizing semantic features from the pre-trained DINO v3 model \cite{voDINOv32025} to localize edited regions based on semantic similarity.

Specifically, as shown in Fig \ref{fig:feature_extraction}, we extract dense features from both the aligned original and edited images using the pre-trained DINO v3 Large model. We then compute the cosine similarity between features at corresponding spatial locations. To automate mask generation, we employ Otsu's method on the similarity map to dynamically determine an optimal threshold. Regions with similarity scores below this threshold are classified as edited, while those above are considered non-edited. The detailed mask annotation pipeline is provided in Algorithm 2 of the Supplementary Material.

\section{PromptForge-350k Dataset}
\label{sec:dataset}

In this section, we first provide additional details regarding our dataset construction process. We then analyze the computational cost of the proposed annotation framework, and finally evaluate the annotation quality through a user study.

\subsection{Dataset Construction}

We applied our proposed annotation pipeline to process image pairs from the X2Edit dataset~\cite{maX2EditRevisitingArbitraryInstruction2025} (comprising subsets edited by BAGEL~\cite{dengEmergingPropertiesUnified2025b}, Kontext~\cite{labsFLUX1KontextFlow2025a}, and Step1x~\cite{liuStep1XEditPracticalFramework2025a}) and the PicoBanana dataset~\cite{qianPicoBanana400KLargeScaleDataset2025} (edited by Nano-Banana~\cite{comaniciGemini25Pushing2025}). Given our focus on local editing, we filtered the datasets based on their ``editing task'' metadata, excluding global editing tasks such as style transfer, color tone conversion, and viewpoint changes. The final dataset includes eight distinct editing task types: object replacement, addition, removal, material conversion, pose change, text editing, portrait editing, and object state conversion. The left panel of Fig \ref{fig:time_breakdown} illustrates the distribution of these editing tasks.

To ensure dataset quality, we discarded samples meeting any of the following criteria: (1) Fewer than 10 keypoints extracted from either image, (2) Fewer than 10 matched keypoint pairs, or (3) An inlier ratio below 60\%. Such samples typically correspond to cases where the edited region is excessively large, the image content is overly simple, or the editing operation failed.

In total, the constructed dataset comprises 354,258 pairs, including 100,000 pairs from each of the three X2Edit subsets and 54,258 pairs from Picobanana. The dataset was randomly partitioned into training and test sets with a 95:5 split. Image resolutions range from 512×512 to 1024×1024, with edited region masks standardized to 128×128.

\subsection{Computational Cost Analysis}

\begin{figure}[htbp]
    \centering
    \includegraphics[width=1.0\linewidth]{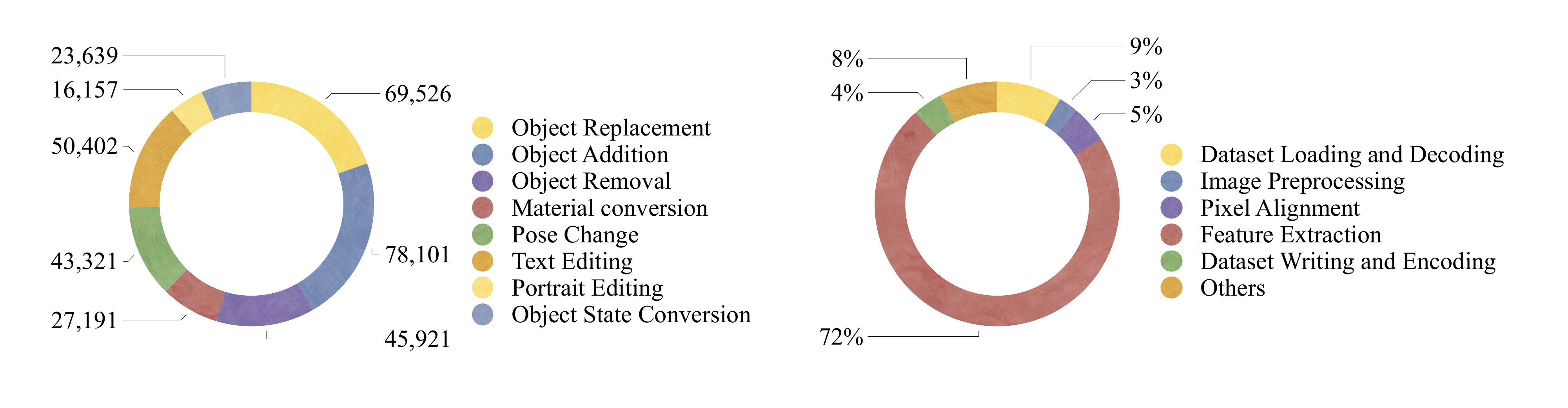}
    \caption{Statistics of the proposed dataset. Left: Distribution of editing task categories. Right: Time consumption breakdown of each operation during the dataset construction process.}
    \label{fig:time_breakdown}
\end{figure}

 Dataset construction was performed on a server equipped with an Intel Xeon Platinum 8469C CPU and 8 NVIDIA H20 GPUs, taking approximately 103 hours. As illustrated in the right panel of Fig \ref{fig:time_breakdown}, we profiled the computational cost of each stage. Notably, our proposed pixel alignment method incurs a marginal overhead of 5\%, while feature extraction accounts for 72\% of the total runtime. To further enhance efficiency, employing semantic feature extraction models with faster inference speeds than DINO v3 Large could effectively accelerate the dataset construction process.

\subsection{Dataset Quality Assessment}
\label{subsec:quality_assessment}
To assess the quality of the dataset, we compiled a set of 200 samples by randomly selecting 50 samples from each of the four subsets. Three volunteers were recruited to rate the annotation quality by assigning each sample to one of the following three categories: (1) Perfect: The annotation is precise or contains only negligible noise; (2) Minor Error: The annotation is largely accurate, with erroneous areas covering less than 20\% of the edited region; and (3) Significnt Error: The annotation contains substantial errors. Discrepancies among the volunteers were resolved via majority voting. The final classification results were 172, 20, and 8, respectively. These results demonstrate that our annotation method yields reliable masks for the edited regions in the vast majority of cases.
\section{ICL-Net}
\label{sec:forge_localization}

\begin{figure}[hbtp]
    \centering
    \includegraphics[width=1.0\linewidth]{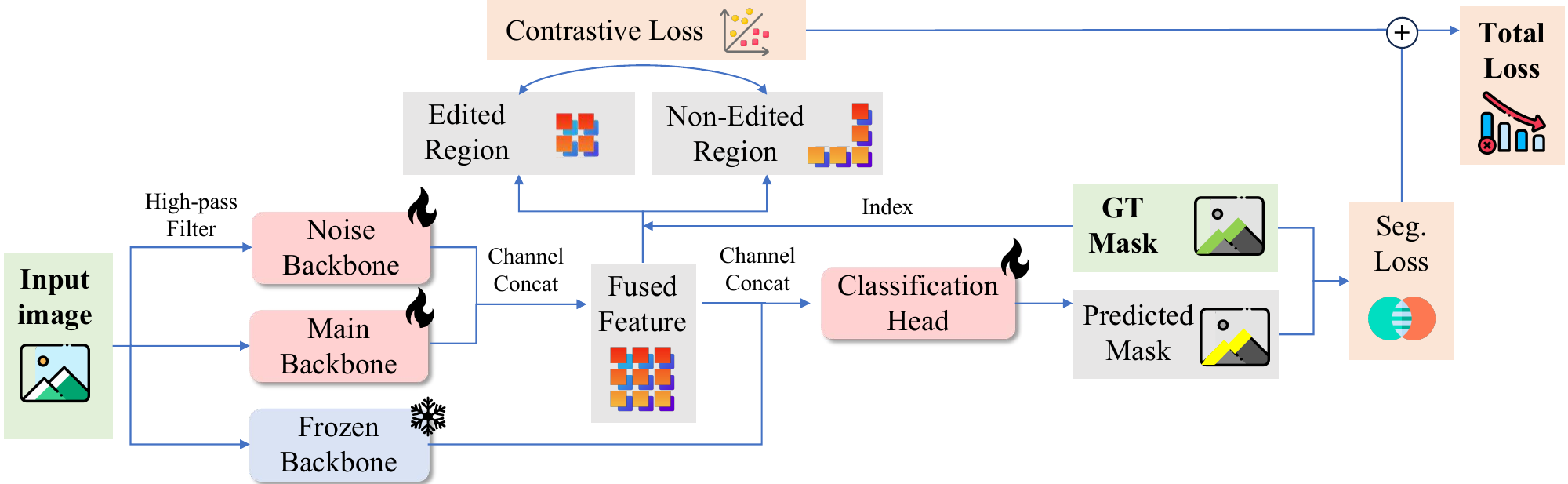}
    \caption{Architecture of the proposed forgery localization network, ICL-Net. The network features three parallel backbones and is optimized via intra-image contrastive loss and segmentation loss.}
    \label{fig:network_structure}
\end{figure}


Unlike traditional Photoshop editing or mask-based AI editing, which operate on specific local regions, prompt-based editing involves a global re-synthesis process where the entire image is regenerated by the model. Consequently, the forensic traces relied upon by traditional forgery localization methods (such as JPEG compression artifacts, splicing boundaries, upsampling anomalies, and CMOS sensor noise) are significantly diminished or obliterated during this process. This necessitates the development of forgery localization methods that are independent of these specific priors.

We posit that in prompt-based editing, non-edited regions structurally and semantically preserve the original image content, whereas edited regions are synthesized entirely based on editing prompts and context. This discrepancy leads to subtle inconsistencies in textural details and semantic distributions between the edited and non-edited regions. Based on this insight, as illustrated in Fig \ref{fig:network_structure}, we introduce ICL-Net, a forgery localization network based on \textbf{I}ntra-image \textbf{C}ontrastive \textbf{L}earning, which guides the network to learn discriminative forgery features. ICL-Net enables precise localization of manipulated regions despite the absence of traditional forensic traces.

\subsection{Network Structure}

\textbf{\ \ \ \ Triple-Stream Backbone.} ICL-Net comprises three parallel backbones with identical structures, all initialized with the same pre-trained SegFormer-B4 \cite{xieSegFormerSimpleEfficient2021} model. Specifically, the Noise Backbone focuses on processing the high-frequency components of the input image, aiming to capture fine-grained textural artifacts associated with forgery. The Main Backbone focuses on extracting semantic-level forgery features. To preserve the generalizable knowledge learned from pre-training and mitigate catastrophic forgetting, we incorporate a Frozen Backbone, which shares the same initial parameters as the Main Backbone but keeps its parameters fixed during training.

 \textbf{Classification Head.}
  Features processed by the three backbones are concatenated along the channel dimension and fed into a trainable Classification Head. It consists of four groups of convolutional and upsampling layers, designed to increase the spatial resolution to match the ground truth mask while reducing the channel dimension to 1, ultimately generating the predicted mask.

\subsection{Loss Function}
\textbf{\ \ \ \ Intra-Image Contrastive Loss.}
Features maps from the Noise and Main backbones are concatenated along the channel dimension to yield a fused feature. Guided by the ground truth mask, we spatially partition these fused features into two distinct sets: those corresponding to forged (edited) regions and those corresponding to real (non-edited) regions. Subsequently, a contrastive loss is computed to maximize the separability between these two feature categories. The formulation is as follows:

\begin{equation}
\label{eq:contrastive_loss}
    \mathcal{L}_{\text{contrastive}} = -\frac{1}{n_f}\sum_{i=1}^{n_f}\log \left( \frac{\frac{1}{n_f} \sum_{j=1}^{n_f}\exp(\operatorname{sim}(F_f[i], F_f[j]) / \tau)}{\sum_{k=1}^{n_r} \exp(\operatorname{sim}(F_f[i],F_r[k]) / \tau)} \right)
\end{equation}

In Equation \ref{eq:contrastive_loss}, $F_f$ and $F_r$ denote the set of feature vectors corresponding to the forgery and real regions, respectively, while $n_f$ and $n_r$ represent the number of pixels in these regions. To balance computational cost and training efficiency, we randomly sample a subset of pixels if the number of candidates exceeds a predifined threshold during training. Additionally, $\text{sim}(\cdot)$ denotes cosine similarity, and $\tau$ is the temperature parameter.

  \textbf{Segmentation Loss.}
The forgery localization task inherently suffers from severe class imbalance, as forged regions typically occupying less than 20\% of the total image area. Relying solely on the Binary Cross-Entropy (BCE) loss commonly used in semantic segmentation can lead to suboptimal convergence. To address this, we employ a hybrid segmentation loss combining Focal Loss and Dice Loss. This combination effectively mitigates class imbalance and compels the network to focus on hard-to-classify classes, thereby enhancing training stability.

Finally, the Contrastive Loss and Segmentation Loss are aggregated to form the Total Loss, which guides the network optimization. Let $\hat{M}$ and $M$ denote the predicted mask and ground truth mask, respectively:

\begin{equation}
    \mathcal{L}_{\text{total}} = \lambda_1 \mathcal{L}_{\text{contrastive}} + 
                                 \lambda_2 \mathcal{L}_{\text{Dice}}(\hat{M}, M) + 
                                 \lambda_3 \mathcal{L}_{\text{Focal}}(\hat{M}, M)
    \label{eq:total_loss}
\end{equation}

\section{Experiment}
\label{sec:experiment}
\subsection{Experimental Setup}

\textbf{\ \ \ \ Parameters.} We resize input images to $512\times 512$ and apply random JPEG compression and cropping for training data augmentation. For the objective function, the loss weights are configured as $\lambda_1=1, \lambda_2=4, \lambda_3=20$. Training is conducted using the AdamW optimizer with an initial learning rate of $1e^{-4}$. To ensure stable convergence, the learning rate is halved if the validation performance plateaus for three consecutive evaluations (performed every 800 iterations), with a minimum lower bound of $1e^{-8}$.

  \textbf{Metrics.} We employ pixel-level F1 score, IoU (Intersection over Union), Precision, and Recall as evaluation metrics. Consistent with common practice, we report the F1 score for the forged class to focus specifically on forged regions. This is crucial because, due to significant class imbalance, a random guess would only achieve an F1 score of approximately 15\%.

  \textbf{Comparision Methods.} We benchmark our ICL-Net against five state-of-the-art forgery localization methods: MVSS-Net (PAMI 2022)\cite{dongMVSSNetMultiViewMultiScale2023}, TruFor (CVPR 2023)\cite{guillaroTruForLeveragingAllround2023}, FOCAL (TDSC 2025) \cite{wuRethinkingImageForgery2025}, Mesorch (AAAI 2025)\cite{zhuMesoscopicInsightsOrchestrating}, and NFA-ViT (AAAI 2026)\cite{caiZoomingFakesNovel2025a}. To ensure a fair comparison, we retrain all baseline methods on our dataset, strictly adhering to the training protocols and hyperparameters detailed in their respective papers.

\subsection{Main Results}


\begin{table*}[t]
  \centering
  \caption{Performance comparison between ICL-Net and SOTA methods. Our method achieves the best performance on all four subsets. The best and second-best results are highlighted in bold and underlined, respectively.}
  \label{tab:model_comparison}
  
  \definecolor{lightgray}{rgb}{0.95, 0.95, 0.95}
  \renewcommand{\arraystretch}{1.0} 
  
  \newcolumntype{Y}{>{\centering\arraybackslash}X}
  
  \begin{tabularx}{\textwidth}{l p{2.8cm} YYYYY}
    \toprule
    \multirow{2}{*}{\textbf{Metric}} & \multirow{2}{*}{\textbf{Method}} & \multicolumn{5}{c}{\textbf{Subset of PromptForge-350K}} \\ 
    \cmidrule(lr){3-7} 
    & &Nano. & BAGEL & Kontext & Step1x & Average \\ 
    \midrule

    \multirow{6}{*}{\textbf{F1}}
    & MVSS-Net   & 31.81 & 44.19 & 54.26 & 52.61 & 45.72 \\
    & TruFor     & 45.30 & 54.58 & 70.26 & 61.50 & 58.61 \\
    & FOCAL      & {\ul 59.28} & 55.49 & {\ul 81.11} & 71.09 & {\ul 66.74} \\
    & Mesorch    & 50.18 & 53.18 & 74.91 & 66.12 & 61.10 \\
    & NFA-ViT    & 54.84 & {\ul 61.76} & 77.19 & {\ul 72.30} & 66.52 \\
    & \cellcolor{lightgray} ICL-Net (ours)  & \cellcolor{lightgray} \textbf{65.75} & \cellcolor{lightgray} \textbf{70.15} & \cellcolor{lightgray} \textbf{84.26} & \cellcolor{lightgray} \textbf{80.62} & \cellcolor{lightgray} \textbf{75.20} \\
    \midrule

    \multirow{6}{*}{\textbf{IoU}}
    & MVSS-Net   & 22.42 & 33.02 & 42.28 & 41.22 & 34.74 \\
    & TruFor     & 35.41 & 43.72 & 59.66 & 51.80 & 47.65 \\
    & FOCAL      & 44.25 & 40.25 & {\ul 68.70} & 57.32 & 52.63 \\
    & Mesorch    & 41.54 & 44.11 & 65.19 & 57.29 & 52.03 \\
    & NFA-ViT    & {\ul 46.07} & {\ul 52.19} & 67.83 & {\ul 63.46} & {\ul 57.39} \\
    & \cellcolor{lightgray} ICL-Net (ours)  & \cellcolor{lightgray} \textbf{51.43} & \cellcolor{lightgray} \textbf{55.18} & \cellcolor{lightgray} \textbf{74.23} & \cellcolor{lightgray} \textbf{68.97} & \cellcolor{lightgray} \textbf{62.45} \\
    \midrule

    \multirow{6}{*}{\textbf{Precision}}
    & MVSS-Net   & 32.68 & 42.85 & 49.57 & 51.14 & 44.06 \\
    & TruFor     & 46.82 & 54.52 & 68.68 & 63.18 & 58.30 \\
    & FOCAL      & {\ul 67.96} & 65.83 & {\ul 80.99} & 75.26 & 72.51 \\
    & Mesorch    & 61.20 & 67.32 & 79.63 & 76.77 & 71.23 \\
    & NFA-ViT    & 64.38 & {\ul 69.27} & 80.71 & {\ul 77.40} & {\ul 72.94} \\
    & \cellcolor{lightgray} ICL-Net (ours)  & \cellcolor{lightgray} \textbf{69.27} & \cellcolor{lightgray} \textbf{73.77} & \cellcolor{lightgray} \textbf{83.27} & \cellcolor{lightgray} \textbf{81.45} & \cellcolor{lightgray} \textbf{76.94} \\
    \midrule

    \multirow{6}{*}{\textbf{Recall}}
    & MVSS-Net   & 44.38 & 58.19 & 71.82 & 64.55 & 59.74 \\
    & TruFor     & 53.80 & {\ul 63.78} & 77.66 & 66.30 & 65.39 \\
    & FOCAL      & 53.53 & 49.01 & {\ul 81.47} & 68.12 & 63.03 \\
    & Mesorch    & 49.66 & 50.44 & 74.70 & 63.84 & 59.66 \\
    & NFA-ViT    & {\ul 54.71} & 61.99 & 76.98 & {\ul 71.77} & {\ul 66.36} \\
    & \cellcolor{lightgray} ICL-Net (ours)  & \cellcolor{lightgray} \textbf{63.50} & \cellcolor{lightgray} \textbf{69.30} & \cellcolor{lightgray} \textbf{85.71} & \cellcolor{lightgray} \textbf{80.02} & \cellcolor{lightgray} \textbf{74.63} \\
    \bottomrule
  \end{tabularx}
\end{table*}

\begin{figure}
    \centering
    \includegraphics[width=1.0\linewidth]{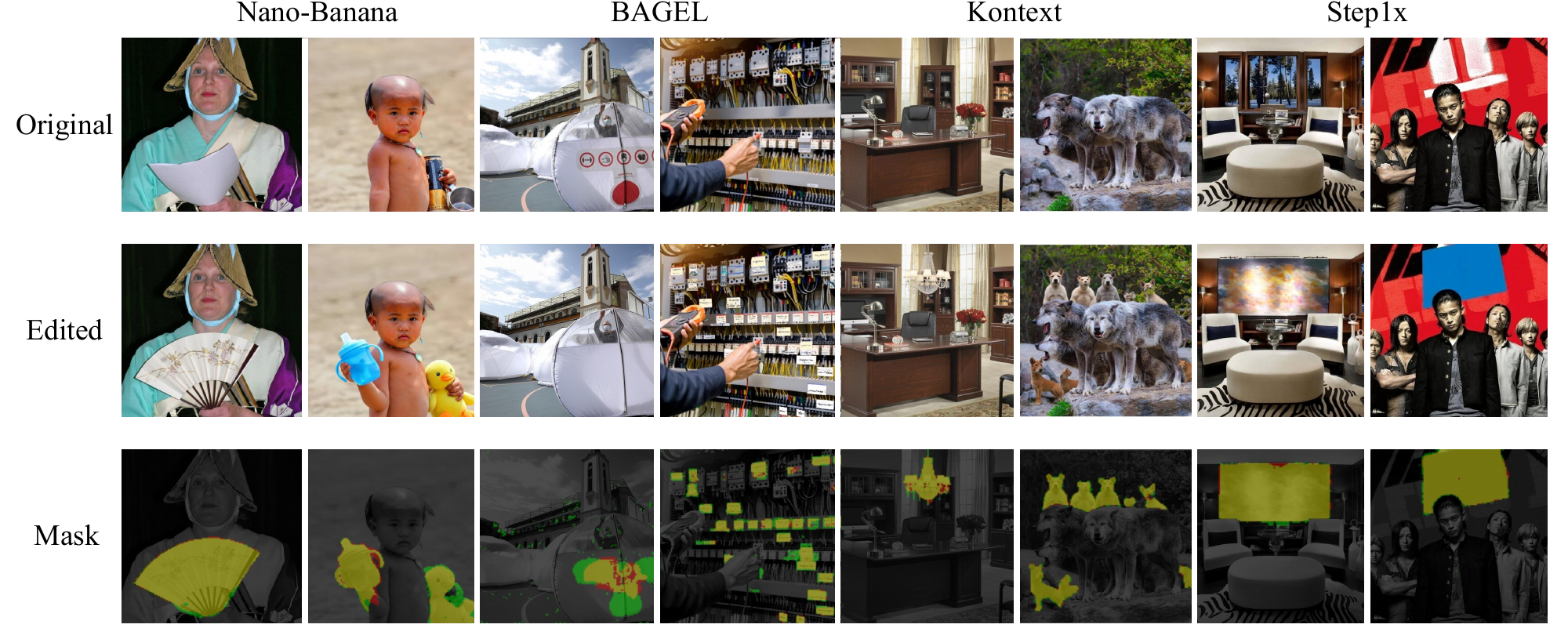}
    \caption{Subjective evaluation results of ICL-Net. The third row highlights the predicted forged regions in red, the ground-truth edited regions in green, and the overlap between predictions and ground truth in yellow.}
    \label{fig:subjective_eval}
\end{figure}

As presented in Tab \ref{tab:model_comparison}, ICL-Net outperforms all competing methods, achieving an average pixel-level F1 score of 75.20\% and an IoU of 62.45\%. 

 \textbf{Performance on Different Subsets.} Among the four editing methods, the closed-source Nano-Banana subset poses the greatest challenge. We attribute this to the fact that the network architectures and training data of closed-source models likely diverge significantly from open-source alternatives, resulting in distinct and more elusive forgery traces.

 \textbf{Analysis of Competitors.} Among the comparison methods, NFA-ViT, Mesorch, and FOCAL exhibit relatively compatitive performance. We reason that while these methods focus on enhancing high-frequency forgery traces, they do not explicitly suppress semantics features. In contrast, MVSS-Net and TruFor employ mechanisms specifically designed to decouple semantic information to avoid interference. While this strategy enhances generalization for traditional splicing or copy-move forgery, it proves detrimental for prompt-based editing localization, where semantic inconsistencies are crucial cues.

 \textbf{Qualitative Analysis.} Qualitative comparisons are visualized in Fig \ref{fig:subjective_eval}. The first two rows display the original and edited images, respectively. The third row illustrates the overlay of predicted masks (red) and ground truth masks (green), with the intersection (yellow) indicating correct predictions. These visualizations qualitatively validate the superior precision of our proposed method.

\subsection{Robustness Analysis}

\begin{table}[t]
  \centering
  \caption{Robustness analysis. We evaluate ICL-Net under varying degrees of JPEG compression and random cropping.}
  \label{tab:robustness}
  
  \definecolor{lightgray}{rgb}{0.95, 0.95, 0.95}
  \renewcommand{\arraystretch}{1.1} 
  
  \begin{tabularx}{\textwidth}{p{3.8cm}cCCCC}
    \toprule
    \textbf{Perturbation} & \textbf{Factor} & \textbf{F1} & \textbf{IoU} & \textbf{Precision} & \textbf{Recall} \\
    \midrule

    \rowcolor{lightgray}
    ICL-Net (baseline) & None & 75.20 & 62.45 & 76.94 & 74.13 \\
    \midrule
    
    \multirow{4}{=}{ICL-Net w/ \\ JPEG Compression} 
      & 90 & 74.83 & 61.92 & 74.74 & 75.68 \\
      & 80 & 74.45 & 61.37 & 74.89 & 74.60 \\
      & 70 & 73.49 & 60.23 & 75.26 & 72.29 \\
      & 60 & 72.24 & 58.82 & 75.73 & 69.58 \\
    \midrule
    
    \multirow{4}{=}{ICL-Net w/ \\ Random Crop} 
      & 10\% & 76.82 & 64.17 & 77.48 & 76.60 \\
      & 20\% & 76.49 & 63.67 & 76.99 & 76.45 \\
      & 30\% & 74.68 & 61.35 & 73.81 & 76.22 \\
      & 40\% & 69.22 & 55.11 & 68.88 & 71.06 \\
    \midrule

    \rowcolor{lightgray}
    {ICL-Net w/ JPEG+Crop} 
      & \textbf{J80 + C20\%} & \textbf{74.61} & \textbf{61.56} & \textbf{74.93} & \textbf{75.06} \\
    \bottomrule
  \end{tabularx}
\end{table}

To evaluate the ICL-Net's robustness against real-world degradations, we applied JPEG compression and random cropping to the test set, to simulate artifacts introduced during social media dissemination. As quantitative results in Tab \ref{tab:robustness} demonstrate, under simultaneous perturbations of JPEG compression (quality factor 80) and 20\% cropping, the F1 score and IoU of our network decreased by a mere 0.58\% and 0.86\%, respectively. This indicates that our method does not heavily rely on fragile forgery traces susceptible to being distorted by image degradations, highlighting its promising potential for real-world social network applications.

\subsection{Cross-Model Generalization}

\begin{table}[!htbp]
  \centering
  \caption{Cross-model generalization performance. We exclude one subset from the training set and evaluate the network's ability to generalize to unseen editing methods.}
  \label{tab:cross_model}
  
  \definecolor{lightgray}{rgb}{0.95, 0.95, 0.95}
  \definecolor{mediumgray}{rgb}{0.9, 0.9, 0.9}
  
  \renewcommand{\arraystretch}{1.1}
  
  \begin{tabularx}{\textwidth}{p{2.2cm}p{2.4cm}CCCC}
    \toprule
    \textbf{Subset} & \textbf{Setting} & \textbf{F1} & \textbf{IoU} & \textbf{Precision} & \textbf{Recall} \\
    \midrule
    
    & In-domain   & 65.75 & 51.43 & 69.27 & 63.50 \\
    \multirow{-2}{=}{Nano-Banana} & \cellcolor{lightgray} Out-of-domain   & \cellcolor{lightgray} 31.07 & \cellcolor{lightgray} 18.95 & \cellcolor{lightgray} 35.25 & \cellcolor{lightgray} 29.49 \\
    \midrule
    
    & In-domain   & 70.15 & 55.18 & 73.77 & 67.30 \\
    \multirow{-2}{=}{BAGEL} & \cellcolor{lightgray} Out-of-domain   & \cellcolor{lightgray} 51.07 & \cellcolor{lightgray} 36.39 & \cellcolor{lightgray} 64.67 & \cellcolor{lightgray} 43.63 \\
    \midrule

    & In-domain   & 84.26 & 74.23 & 83.27 & 85.71 \\
    \multirow{-2}{=}{Kontext} & \cellcolor{lightgray} Out-of-domain   & \cellcolor{lightgray} 68.44 & \cellcolor{lightgray} 53.53 & \cellcolor{lightgray} 75.10 & \cellcolor{lightgray} 63.93 \\
    \midrule

    & In-domain   & 80.62 & 68.97 & 81.45 & 80.02 \\
    \multirow{-2}{=}{Step1x} & \cellcolor{lightgray} Out-of-domain   & \cellcolor{lightgray} 70.93 & \cellcolor{lightgray} 57.12 & \cellcolor{lightgray} 77.82 & \cellcolor{lightgray} 65.41 \\
    \midrule
    
    & \cellcolor{mediumgray} In-domain   & \cellcolor{mediumgray} 75.20 & \cellcolor{mediumgray} 62.45 & \cellcolor{mediumgray} 76.94 & \cellcolor{mediumgray} 74.13 \\
    \multirow{-2}{=}{\textbf{Average}} & \cellcolor{mediumgray} \textbf{Out-of-domain}   & \cellcolor{mediumgray} \textbf{55.37} & \cellcolor{mediumgray} \textbf{41.50} & \cellcolor{mediumgray} \textbf{63.21} & \cellcolor{mediumgray} \textbf{50.62} \\
    
    \bottomrule
  \end{tabularx}
\end{table}

To assess cross-model generalization, we adopted a Leave-One-Out (LOO) strategy, where one subset was excluded from training and subsequently used for evaluation. As illustrated in Tab \ref{tab:cross_model}, performance declines across all unseen editing methods. A notable trend is that the drop in Recall is more pronounced than in Precision, suggesting that ICL-Net tends to yield conservative predictions when encountering forged images from unseen models.

For the three open-source editing models, our approach achieves an IoU ranging from 36.39\% to 57.12\%, even though they were unseen during the training phase. However, for Nano-Banana, ICL-Net attains an IoU of only 18.95\%, significantly lower than the generalization performance observed on the three open-source counterparts. This indicates that the forgery artifacts of closed-source methods differ significantly from those of open-source methods. Consequently, forgery localization networks trained exclusively on open-source models exhibit limitations when generalizing to closed-source methods.

\subsection{Ablation Studies}

We conducted ablation studies to validate the effectiveness of our key design components: the contrastive loss, the noise backbone and the frozen backbone. The results presented in Tab \ref{tab:ablation} demonstrate that each component plays a critical role. In particular, the exclusion of the contrastive loss leads to a significant performance drop, with F1 and IoU declining by 15.6\% and 17.38\%, respectively. This indicates that the contrastive loss is instrumental in guiding the network to differentiate between edited and non-edited regions, thereby ensuring the network learns discriminative features.

\begin{table}[thb]
  \centering
  \caption{Ablation study: removing the contrastive loss, noise backbone, or the frozen backbone each leads to a noticeable drop in localization performance.}
  \label{tab:ablation}
  
  \definecolor{lightgray}{rgb}{0.95, 0.95, 0.95}
  
  \renewcommand{\arraystretch}{1.1}
  
  \begin{tabularx}{\textwidth}{p{5.5cm}CCCC}
    \toprule
    \textbf{Operation} & \textbf{F1} & \textbf{IoU} & \textbf{Precision} & \textbf{Recall} \\
    \midrule
    
    \rowcolor{lightgray} 
    ICL-Net (baseline) & \textbf{75.20} & \textbf{62.45} & \textbf{76.94} & 74.13 \\
    
    ICL-Net w/o contrastive loss & 59.63 & 45.07 & 60.45 & 59.55 \\
    ICL-Net w/o noise backbone   & 73.21 & 59.84 & 71.57 & \textbf{75.43} \\
    ICL-Net w/o frozen backbone  & 70.44 & 56.95 & 75.61 & 66.81 \\
    
    \bottomrule
  \end{tabularx}
\end{table}

\section{Limitations and Future Work}
\label{sec:discussion}

 \textbf{\ \ \ \ Limitations of the Annotation Pipeline.} As discussed in Section \ref{subsec:quality_assessment}, our dataset annotation method yields incorrect masks in a small fraction of edge cases. Manual inspection reveals that these mislabeled samples are primarily concentrated in scenarios involving solely object color modifications, such as changing an apple from red to green. This is likely attributed to the DINO v3 model's relatively low sensitivity to object color, as it prioritizes semantic information. Future work could address this by incorporating additional discriminative cues for joint decision-making or by fine-tuning the DINO v3 model to further enhance annotation accuracy.

 \textbf{Generalization Gap.} Although our forgery localization method demonstrates satisfactory performance, its generalization capability to unseen closed-source models remains limited. Enhancing cross-model generalization, especially for closed-source models, represents a critical direction for future research.




\section{Conclusion}
\label{sec:conclusion}
In this work, we address the critical yet underexplored challenge of image forgery localization for prompt-based AI editing. We present PromptForge-350k, a large-scale dataset constructed via our fully automated annotating pipeline. Furthermore, we propose ICL-Net, an effective forgery localization network that incorporates intra-image contrastive learning. Extensive experiments validate the superiority of our approach, which achieves state-of-the-art performance on PromptForge-350k while demonstrating resilience against common post-processing perturbations. We hope that PromptForge-350k and ICL-Net will serve as a solid foundation to facilitate future research in the growing field of prompt-based AI image forgery localization.

\bibliographystyle{plain}
\bibliography{main}

\end{document}